\documentclass[10pt,twocolumn,letterpaper]{article}

\usepackage{cvpr}              % To produce the CAMERA-READY version
\definecolor{cvprblue}{rgb}{0.21,0.49,0.74}
\usepackage[colorlinks,linkcolor=red,anchorcolor=green,citecolor=cvprblue]{hyperref}
\usepackage[accsupp]{axessibility}  % Improves PDF readability for those with disabilities.

\usepackage{algorithm}
\usepackage{algorithmic}
\usepackage{booktabs}
\usepackage{multirow}
\usepackage{amsmath}
\usepackage{amssymb}
\usepackage{colortbl}
\usepackage{xcolor}
\usepackage{graphicx}
\usepackage{subfloat}
\usepackage{multicol}
\usepackage{xspace}

\def\paperID{1955} % *** Enter the Paper ID here
\def\confName{CVPR}
\def\confYear{2025}

\title{Image Over Text: Transforming Formula Recognition Evaluation with Character Detection Matching}

\def\spaces{~~~~~~}
\author{
Bin Wang\textsuperscript{1}\thanks{Equal contribution.~~\textsuperscript{\dag}Corresponding author: heconghui@pjlab.org.cn} \spaces{}  
Fan Wu\textsuperscript{1}\footnotemark[1]\spaces{} 
Linke Ouyang\textsuperscript{1}\footnotemark[1] \spaces{} 
Zhuangcheng Gu\textsuperscript{1} \spaces{} \\
Rui Zhang\textsuperscript{1} \spaces{} 
Renqiu Xia\textsuperscript{1,2}\spaces{} 
Botian Shi\textsuperscript{1}\spaces{} 
Bo Zhang\textsuperscript{1}\spaces{} 
Conghui He\textsuperscript{1}$^{\dagger}$ \\\\
\textsuperscript{1}Shanghai Artificial Intelligence Laboratory\spaces{} \textsuperscript{2}Shanghai Jiao Tong University \\
}

\begin{document}
\maketitle
\begin{abstract}
Formula recognition presents significant challenges due to the complicated structure and varied notation of mathematical expressions. Despite continuous advancements in formula recognition models, the evaluation metrics employed by these models, such as BLEU and Edit Distance, still exhibit notable limitations. They overlook the fact that the same formula has diverse representations and is highly sensitive to the distribution of training data, thereby causing unfairness in formula recognition evaluation. To this end, we propose a Character Detection Matching (CDM) metric, ensuring the evaluation objectivity by designing an image-level rather than a LaTeX-level metric score. Specifically, CDM renders both the model-predicted LaTeX and the ground-truth LaTeX formulas into image-formatted formulas, then employs visual feature extraction and localization techniques for precise character-level matching, incorporating spatial position information. Such a spatially-aware and character-matching method offers a more accurate and equitable evaluation compared with previous BLEU and Edit Distance metrics that rely solely on text-based character matching. Experimentally, we evaluated various formula recognition models using CDM, BLEU, and ExpRate metrics. Their results demonstrate that the CDM aligns more closely with human evaluation standards and provides a fairer comparison across different models by eliminating discrepancies caused by diverse formula representations. Code is available at \url{https://github.com/opendatalab/UniMERNet/tree/main/cdm}.
\end{abstract} 

\begin{figure*}[ht]
\vspace{-2pt}
\begin{center}
    \includegraphics[width=1.0 \linewidth]{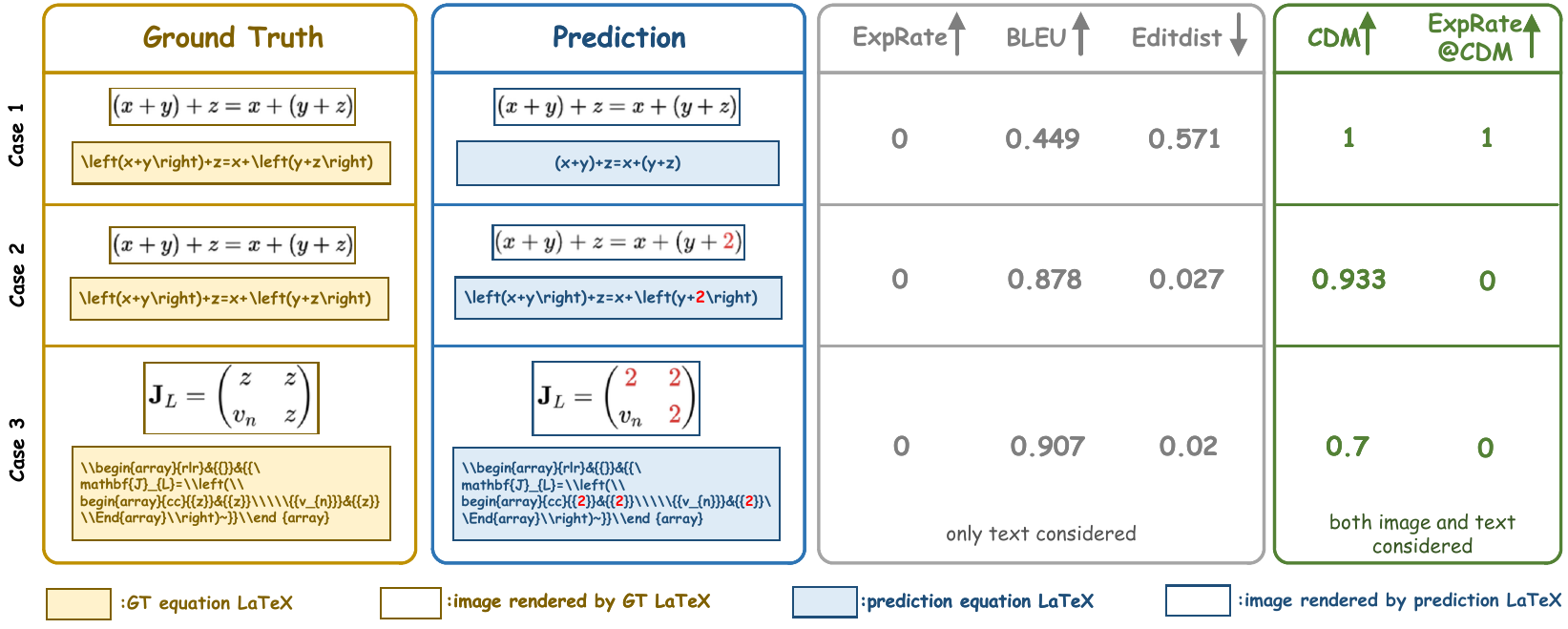}
\end{center}
    \vspace{-12pt}
    \caption{Illustration of the limitations of the existing metrics (ExpRate~\cite{deng2017image}, BLEU~\cite{papineni2002bleu}, and Edit Distance~\cite{levenshtein1966binary}) and the advantages of the proposed CDM. \textbf{Case 1:} Due to different expression styles, the model prediction and Ground Truth (GT) may appear visually similar, but they receive low scores when evaluated using traditional metrics such as BLEU. \textbf{Case 2:} Errors in numeric predictions within formulas are essentially prediction errors. However, when evaluated using BLEU, such errors may receive higher metric scores due to the model predicting an expression style close to the GT. \textbf{Case 3:} Model predictions that are visually obviously incorrect may receive high scores using BLEU metric, which is inconsistent with human judgment standards.
    }
\label{fig:fig1}
\vspace{-10pt}
\end{figure*}

\section{Introduction}

Mathematical formula recognition is crucial in document analysis as it directly impacts the scientific rigor and accuracy of the document content~\citep{xia2024docgenome,wang2024mineru,cheng2023m6doc,li2020docbank}. Unlike standard Optical Character Recognition (OCR) technique, formula recognition presents unique challenges. Formulas often encompass multi-level symbols, subscripts, fractions, and other complicated structures, requiring models to comprehend spatial and structural relationships rather than just linear, sequential text~\citep{kukreja2023recent,li2020docbank}. Besides, formulas exhibit representational diversity, meaning that the same formula can be expressed in multiple valid ways.

In recent years, significant advancements in formula recognition~\cite{deng2017image,zhang2020tree,yuan2022syntax,pix2tex2022,texify2023,liao2020mask,yang2019symmetry} have been primarily driven by the continuous development of artificial intelligence technique~\cite{wei2025vary, blecher2023nougat, yu2024texthawk2}. Besides, commercial formula recognition software like Mathpix\footnote[1]{\url{https://mathpix.com/equation-to-latex}\label{fn:1}} and the recently proposed UniMERNet~\cite{wang2024unimernet} model have achieved impressive results in diverse real-world settings. Despite these advancements, the existing evaluation metrics~\cite{papineni2002bleu,levenshtein1966binary} for formula recognition still face some challenges. Commonly-used metrics such as BLEU~\cite{papineni2002bleu} and Edit Distance~\cite{levenshtein1966binary} primarily rely on text-based character matching, which introduces several limitations as follows:

\noindent \textbf{(1) Low Metric Reliability.} BLEU and Edit Distance are reliable for evaluating the quality of text-level similarity. However, the diversity in formula representations makes these text-level evaluation metrics inadequate for precisely reflecting formula recognition quality. For example, as shown in \Cref{fig:fig1} (Case 1), a model's prediction might render an image identical to the ground truth formula. However, due to the variations in formula expression styles, the evaluation results obtained using the ExpRate~\cite{deng2017image}, BLEU, and Edit Distance may be somewhat misleading.

\noindent \textbf{(2) Unfair Model Comparison.} Current metrics may be susceptible to discrepancies between the distributions of training and testing data. As illustrated in \Cref{fig:fig1} (Case 1 and Case 2), a model may produce a correct prediction but score poorly due to representational differences from the ground truth, while an incorrect prediction might score higher if its representation aligns more closely with the test data distribution.

\noindent \textbf{(3) Lack of Intuitive Scoring.} There can be a significant discrepancy between BLEU scores and human perception. For instance, in \Cref{fig:fig1} (Case 3), a model's prediction contains many errors, yet the BLEU score is as high as 0.907, which does not align with human judgment.

To address these issues, we propose a novel evaluation metric for formula recognition: Character Detection Matching (CDM). The proposed CDM regards the formula recognition evaluation as an image-based object detection task, by converting both the predicted LaTeX and the ground-truth LaTeX formulas into the image-formatted formulas and treating each character as an independent target. This approach overcomes the challenges posed by the diverse expression styles of formulas and aligns more closely with human subjective evaluation standards. CDM offers the advantages as follows: \textbf{1) Accuracy and Reliable.} By calculating metrics in the image space, CDM eliminates issues caused by different valid representations of the same formula, directly reflecting recognition accuracy and aligning more closely with human intuitive perception. \textbf{2) Fairness.} CDM removes the high dependency on consistent data distribution between training and evaluation task, allowing for a fair comparison of different models based on their true recognition capabilities. Our contributions can be summarized as follows:

\begin{itemize}
\item We perform a detailed analysis of the existing formula recognition evaluation metrics, highlighting the limitations of ExpRate and BLEU and their unreliability specifically for evaluating formula recognition tasks.
\item We introduce a novel evaluation metric, CDM, which assesses formula recognition quality by performing visual character matching between rendered images of predicted and ground-truth formulas, providing an intuitive and fair evaluation standard.
\item We validate CDM's effectiveness through extensive experiments on various mainstream models and datasets, demonstrating its superiority over traditional metrics like BLEU in assessing formula recognition performance.

\end{itemize}

\section{Related Work}

\subsection{Formula Recognition Algorithms}

Initially, researchers employ specific grammar rules to represent the spatial structure of formulas, including graph grammars~\cite{LavirotteP98}, relational grammars~\cite{maclean2013new}, and probabilistic grammars~\cite{awal2014global,alvaro2016integrated}. Besides, the CROHME competitions~\cite{mouchere2013icdar,le2019pattern,mouchere2014icfhr,mouchere2016icfhr2016,mahdavi2019icdar} have promoted the development of handwritten formula recognition by incorporating deep learning algorithms. Key contributions include a neural encoder-decoder model with coarse-to-fine attention~\cite{deng2017image}, a tree-structured decoder~\cite{zhang2020tree}, and the Counting-Aware Network~\cite{li2022counting}, which integrates a weakly-supervised counting module. The ABM network~\cite{bian2022handwritten} employs mutual distillation and an Attention Aggregation Module, while a transformer-based decoder~\cite{zhao2021handwritten} simplifies model architecture. The Syntax-Aware Network (SAN)~\cite{yuan2022syntax} models recognition as a tree traversal process, significantly improving accuracy for complex expressions. Overall, these models employ ExpRate~\cite{deng2017image} for formula recognition evaluation.

In document information extraction~\cite{xia2024docgenome,roberts2024image2struct,li2022counting,xia2024chartx}, Donut \cite{kim2022ocr} directly converts input documents into structured outputs without using traditional OCR tools. Texify~\cite{texify2023} and UniMERNet~\cite{wang2024unimernet} are designed using Donut~\citep{kim2022ocr}, utilizing more diverse datasets and data augmentation operations. Nougat~\citep{blecher2023nougat} is designed to convert PDF documents from screenshot to Markdown format, making the document content (\textit{e.g.} table and formula) easier to edit. These methods use BLEU~\cite{papineni2002bleu} and Edit Distance~\cite{levenshtein1966binary}  metrics for formula recognition evaluation.

\subsection{Formula Recognition Evaluation Metrics}

\textbf{BLEU} is initially proposed for machine translation tasks, matching standard and machine-translated texts using N-grams (sequences of N words) between the generated and the reference texts. It applies a brevity penalty factor to produce the final BLEU score~\cite{papineni2002bleu}:
\begin{equation}
    BLEU = BP \cdot \exp \left( \sum_{n=1}^{N} w_n \log p_n \right),
\end{equation}
where BP is the brevity penalty factor, and \( p_n \) is the N-gram match result, with \( n \) ranging from 1 to 4. 

\noindent\textbf{Edit Distance} is also commonly-used metric to assess the similarity between the generated and the reference texts. It measures the number of insertions, deletions, or substitutions needed to transform one text into another, with a smaller Edit Distance indicating higher similarity~\cite{levenshtein1966binary}.

\noindent\textbf{ExpRate} refers to the proportion of samples where the texts are exactly matched out of the total number of samples. Compared to BLEU and Edit Distance, ExpRate is coarser and more stringent in evaluation~\cite{li2022counting}.

The above three metrics can effectively evaluate the textual differences between ground truth and reference, making them suitable for tasks requiring strict matches. BLEU and Edit Distance, in particular, provide a finer evaluation of text recognition capabilities compared to ExpRate, making them widely used in extensive text recognition tasks such as document recognition~\cite{blecher2023nougat,huang2024mini}. These metrics are also applied to formula recognition, with most open-source models, such as Pix2Tex~\cite{pix2tex2022} and Texify, adopting them for evaluation and comparison.

In addition to text-based metrics, image edit distance has been explored to measure the accuracy of predicted formulas~\cite{wang2021translating}. Image processing metrics like MSE (Mean Squared Error) and SSIM~\cite{wang2004image} have also been considered. Structuring Chart-oriented Representation Metric (SCRM)~\cite{xia2023structchart} is designed to comprehensively evaluate the information represented by structured triplet representations. However, these metrics are better suited for natural images. For document images such as formula images, even slight character misalignments can result in significant penalties, making these metrics less suitable for formula recognition.

\begin{figure*}[t]
\vspace{-2pt}
\begin{center}
	\includegraphics[width=1.0 \linewidth]{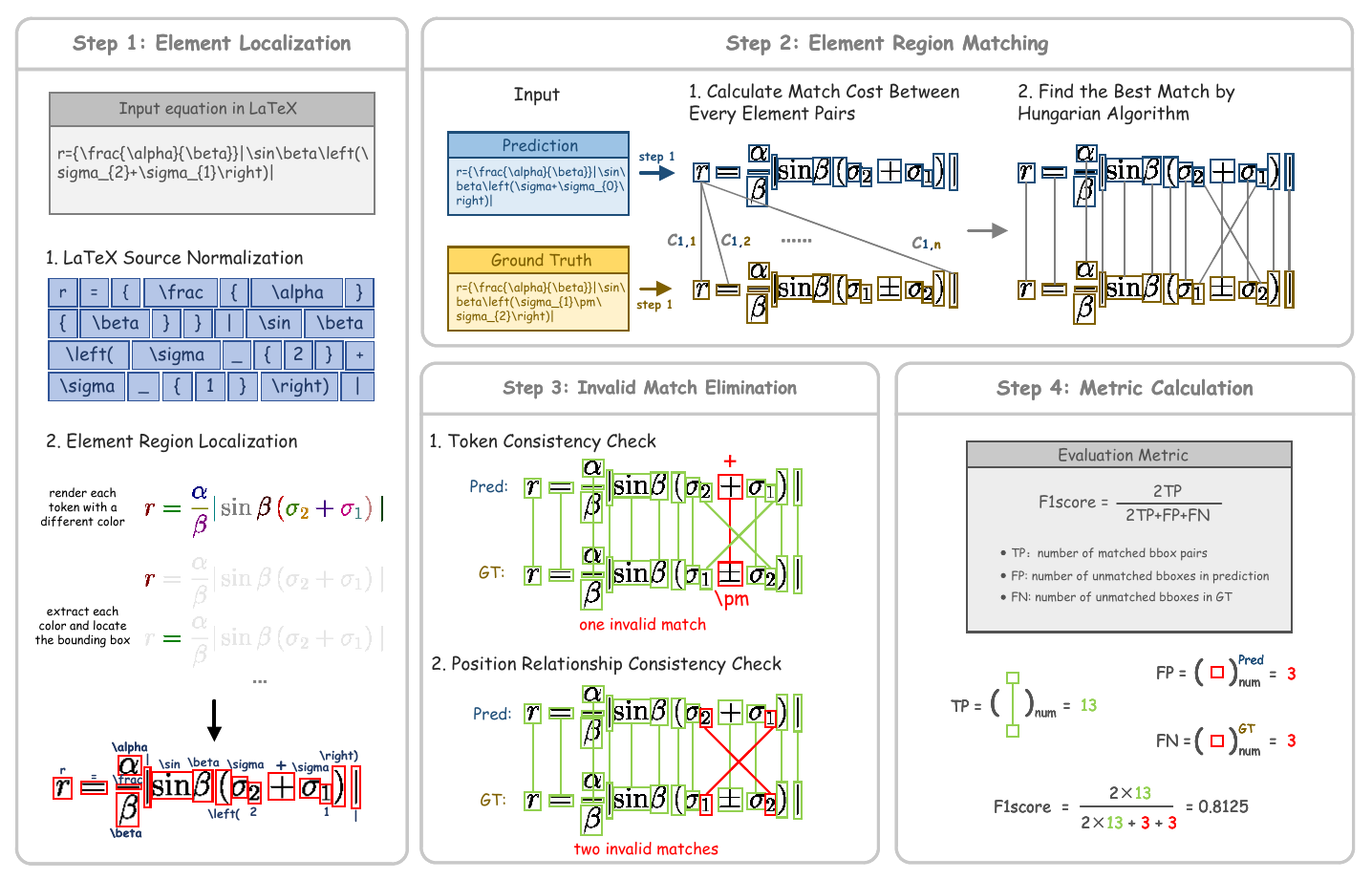}
\end{center}
\vspace{-12pt}
\caption{Overview of the Character Detection Matching (CDM), consisting of four main stages. \textbf{(1)} Element Localization, where bounding boxes of individual elements are extracted. \textbf{(2)} Element Region Matching, which employs a bipartite graph matching method to pair prediction with ground truth elements. \textbf{(3)} Invalid Match Elimination, where inconsistent matches are discarded through token and positional relationship checks. \textbf{(4)} Metric Calculation, where matching accuracy is evaluated using the F1-Score and ExpRate@CDM.}
\label{fig:fig2_CDM}
\vspace{-10pt}
\end{figure*}

\section{Limitations of Current Metrics}

Although ExpRate, BLEU, and Edit Distance are widely used in formula evaluation tasks, they exhibit significant limitations in accurately reflecting formula recognition performance, particularly in scenarios where there are domain gaps between training and testing data distributions. The main reason is that a single formula can have multiple valid LaTeX representations, making the Ground Truth (GT) LaTeX non-unique, which introduces inherent flaws for the formula evaluation.

As illustrated in Case 1 of~\Cref{fig:fig1} earlier, the formula $(x+y)+z=x+(y+z)$ corresponds to the GT annotation \texttt{"\textbackslash{}left(x+y\textbackslash{}right)+z=x+\textbackslash{}left(y+z\textbackslash{}right)"}. When the model's prediction is \texttt{"(x+y)+z=x+(y+z)"}, the prediction is correct because the rendered formula image matches the GT image, despite different LaTeX syntax. Theoretically, the ExpRate/BLEU/Edit Distance results should be 1/1/0, indicating a correct instance. However, in practice, ExpRate is 0, BLEU is 0.449, and Edit Distance is 0.571, failing to accurately assess the formula's quality.

The aforementioned issues make it challenging to objectively evaluate the performance of different formula recognition models. For instance, as illustrated in Case 2 of~\Cref{fig:fig1}, 
one character \texttt{"z"} is misrecognized as \texttt{"2"}. The prediction is incorrect, and the ExpRate, BLEU, and Edit Distance metrics reflect this error. However, when compared to Case 1 where the model prediction is correct, the BLEU and Edit Distance metrics for the incorrect prediction in Case 2 are better than those for the correct prediction in Case 1.

A LaTeX regularization method, which abstracts LaTeX code into a tree structure and standardizes elements, addresses LaTeX syntax diversity~\cite{deng2017image}. Pix2tex~\citep{pix2tex2022}, Texify~\citep{texify2023}, and UniMERNet~\citep{wang2024unimernet} use such regularization method as a preprocessing step before evaluation, which can solve part of the syntax inconsistency issue. For instance, \texttt{"x\^{}b\_a"}, \texttt{"x\^{}\{b\}\_\{a\}"}, and \texttt{"x\_\{a\}\^{}\{b\}"} all compile to $x^b_a$. Directly calculating BLEU scores would not correctly assess the model's prediction quality. Regularized code unifies these into a consistent format, such as always adding curly braces and arranging superscripts before subscripts, contributing to the fairness of subsequent metric calculations. However, regularization does not solve all LaTeX syntax diversity issues. Some symbols have multiple representations, such as \texttt{"\textbackslash{}leq"} and \texttt{"\textbackslash{}le"} both representing $\leq$. Exhaustively listing these representations is challenging due to the huge LaTeX symbol library and many additional symbols provided by extension packages (\textit{e.g.}, amsmath, amssymb).

Overall, while regularization mitigates some issues, it does not fully address the inherent limitations of current metrics in evaluating formula recognition performance. This highlights the need for a more robust and comprehensive evaluation metric that can accurately reflect the quality of formula recognition across diverse representations.

\section{Character Detection Matching}

Due to the diversity of LaTeX expressions, text-based character-matching methods are unreliable for formula recognition evaluation. The basic idea of CDM is to compare the rendered images from LaTex text. If the image rendered from the predicted LaTeX source code matches the image rendered from the ground truth LaTeX source code, the formula is considered entirely correct. However, directly comparing the pixel values of the original and predicted formulas is not ideal. Any error or extra/missing character in the prediction can cause subsequent characters to be mismatched. Additionally, two similar formulas might have different layouts, with one being a single-line formula and the other a multi-line formula due to line breaks. Therefore, a more robust algorithm is needed to calculate the match between the predicted result and the ground truth image.

To this end, we propose a metric that incorporates a bipartite matching step for element-level matching in images, providing a more intuitive assessment. As shown in~\Cref{fig:fig2_CDM}, the algorithm consists of four stages as follows.

\subsection{Element Localization}

First, the bounding boxes (bboxes) of each individual element in the rendered image are extracted, followed by the subsequent steps:

\noindent \textbf{LaTeX Source Normalization.}  LaTeX source codes of both the ground truth and predicted formulas are normalized, breaking them down into individual tokens such as \texttt{"2", "a", "A", "\textbackslash alpha", "\textbackslash sin"}. Composite elements are decomposed into individual characters, \textit{e.g.}, \texttt{"\textbackslash frac ab"} is decomposed into \texttt{"\textbackslash frac \{a\} \{b\}"}.

% \noindent \textbf{Element Region Localization.} Each token in color based on the normalized LaTex source code is rendered. For the element $e$ to be localized, render it using \texttt{"\textbackslash black\{e\}"} while rendering other elements using \texttt{"\textbackslash gray\{}$\bar{e}$\texttt{\}"}. By binarizing the fully rendered formula to extract the bounding box of each element, this process is repeated until all elements are accurately localized.

% \noindent \textbf{Element Region Localization.} To accurately identify the position of each character, our initial step involves rendering all tokens in distinct colors. This process begins by constructing an RGB color list based upon fixed intervals. In the experimental setup, we chose an interval of 15, yielding an RGB color sequence that looks like [(0,0,15), (0,0,30), ..., (255, 255, 255)]. With this approach, we generate $(255/15 + 1)^3 = 5832$ distinct colors, plentiful enough to accommodate an extensive formula.
% Subsequent to formulating the color list, we sequentially assign these colors to every token by applying the LaTeX command \texttt{"\textbackslash mathcolor[RGB]\{r,g,b\}"} to each one. As a result, when the image is rendered, each token is represented by a unique color that is directly associated with it.
% For the next phase, we employ image processing techniques to pinpoint all the pixels that match each color. By analyzing these pixels, we can determine the precise position coordinates for the token corresponding to that particular color.

\noindent \textbf{Element Region Localization.}  To accurately detect character positions, we render each token in a unique color. We construct an RGB color list with a fixed interval of 15, creating a sequence from $(0,0,15)$ to $(255,255,255)$. This generates $(255/15 + 1)^3 = 5832$ distinct colors, sufficient for complex formulas. Each token is assigned a unique color using the LaTeX command \texttt{"\textbackslash mathcolor[RGB]\{r,g,b\}"}. After rendering, we apply image processing techniques to identify pixels matching each color, thereby determining the exact coordinates of each token.

\subsection{Element Region Matching}

In this stage, a bipartite matching method pairs the predicted elements with the corresponding ground truth elements. Based on the element localization, two sets are obtained for each formula: one for the ground truth independent elements $y$ and one for the predicted independent elements $\hat{y}$. The number of independent elements in each set is $N_y$ and $N_{\hat{y}}$, respectively, with $N = \min(N_y, N_{\hat{y}})$ being the number of elements in the smaller set.

To measure the similarity between \( y \) and \( \hat{y} \), we match elements in the two sets by identifying the corresponding ground truth element for each predicted element. We use the bipartite matching Hungarian algorithm~\cite{kuhn1955hungarian}, as described in DETR~\cite{carion2020end}, to find a permutation \( \hat{\sigma} \) that minimizes the total matching cost:
\begin{equation}
\hat{\sigma} = \arg \min_{\sigma \in S_N} \sum_{i=1}^N L_{\text{match}}(y_i, \hat{y}_{\sigma(i)}),
\end{equation}
%    L_{\text{match}}=W_t\times L_t + W_v\times L_v + W_p \times L_p  + W_o \times L_o,
\begin{equation}
    L_{\text{match}}=W_t\times L_t + W_p \times L_p  + W_o \times L_o,
\end{equation}
where the matching cost $L_{match}$ is defined as a weighted sum of three components as introduced as follows:

\begin{itemize}
    \item \textbf{Token Matching Cost \( L_t \)}: This component measures whether the tokens corresponding to two bounding boxes are the same. If they are identical, the cost is 0; if they are different, the cost is 1. For tokens that render identically but are different, such as \texttt{"("}, \texttt{"\textbackslash left("}, and \texttt{"\textbackslash big("}, the cost is 0.05, which can be formulated as follows:
    \begin{equation}
    L_{\text{t}} = 
    \begin{cases} 
    0, & \text{if } t_{i} = \hat{t}_{\sigma(i)}; \\ 
    0.05, & \text{if } t_{i} \approx \hat{t}_{\sigma(i)} ; \\ 
    1, & \text{otherwise}; 
    \end{cases}
    \end{equation}
    where \( \approx \) denotes tokens that differ but render identically.

    \item \textbf{Positional Proximity Cost \( L_p \)}: This component measures the proximity of the two bounding boxes' positions using the L1 norm of their coordinates, which can be formulated as follows:
    \begin{equation}
    L_{\text{p}} = \frac{1}{D_b} \times \|b_i - \hat{b}_{\sigma(i)}\|_1 ,
    \end{equation}
    where \( b = [x_1, y_1, x_2, y_2] \), and \( D_b \) is the dimension of the bounding box coordinates.

    \item \textbf{Order Similarity Cost \( L_o \)}: This measures the similarity of the token order in the original LaTeX source (an approximation of reading order). The order is normalized to the range [0, 1], and the L1 norm can be calculated as follows:
    \begin{equation}
    L_{\text{o}} = \frac{1}{D_o} \times \|o_i - \hat{o}_{\sigma(i)}\|_1 ,
    \end{equation}
    where $o_i$ means order of $i$-th token, with dimension $D_o$.
    \end{itemize}

Overall, the weights \( W_t, W_p, W_o \) are used to balance the contributions of the three components. By employing this comprehensive matching strategy, we ensure a more accurate and robust evaluation of the correspondence between the predicted and ground truth elements, thereby improving the overall assessment of formula recognition quality.

\subsection{Invalid Match Elimination}

After pairing the individual elements of the predicted result with the ground truth using the Hungarian matching algorithm, we need to verify these pairs and eliminate invalid matches. This process involves two steps:

\noindent \textbf{Token Consistency Check.} Check whether the elements in each matched pair are consistent in terms of characters. If they are inconsistent, discard the match.

\noindent \textbf{Position Relationship Consistency Check.} The relative positions of elements in mathematical formulas are crucial. For instance, in the expressions $2^3$ and $3^2$, bipartite matching might pair 2 with 2 and 3 with 3, but their meanings and visual representations are entirely different. Thus, we need to check the consistency of the positional relationships within the matched pairs. We treat each element in the matched pair as a bounding box and analyze their relative positions. Specifically, we assume an affine transformation  between the ground truth and predicted elements:
\begin{equation}
    \hat{b}_{\sigma(i)} = \mathbf{A}(b_i),
\end{equation}
where \(\mathbf{A}\) is the affine transformation matrix. To identify inconsistent match pairs, we detect pairs that do not conform to this transformation relationship. We employ the RANSAC algorithm~\cite{fischler1981random} for this purpose. RANSAC can determine the optimal transformation matrix \(\mathbf{A}\) in the presence of noise. Given that formulas are usually horizontally arranged during rendering, we fix the rotation angle in the transformation matrix to 0, considering only translation and scaling. This approach not only improves the convergence speed of the RANSAC algorithm but also enhances the final matching accuracy.

To account for line-breaking effects in formulas, we perform multiple rounds of RANSAC iterations to ensure that as many matched pairs as possible conform to the transformation relationship. After several iterations, matched pairs that still do not conform to the transformation relationship are considered incorrect and are eliminated.

The above two steps effectively eliminate invalid match pairs, ensuring more accurate final matching results.

\begin{figure*}[ht]
\vspace{-8pt}
    \centering
    \subfloat[CDM distribution on UniMER-Test dataset]{
    \begin{minipage}{0.45\textwidth}
        \centering
        \includegraphics[width=\textwidth]{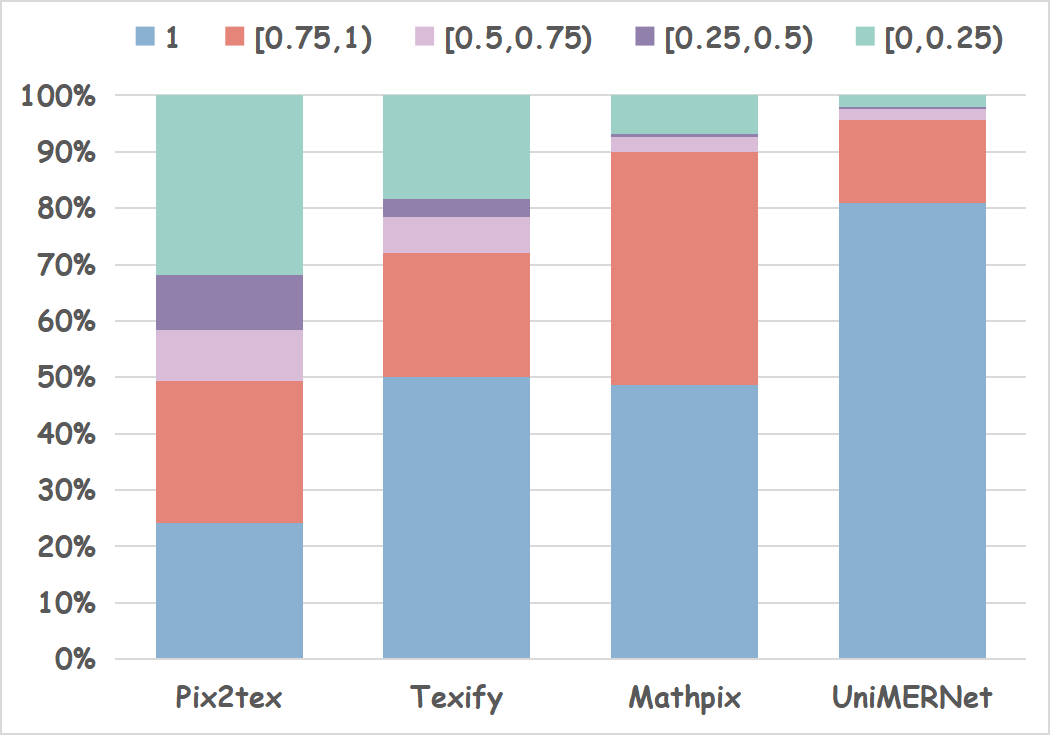}
        \label{fig:fig3_a}
    \end{minipage}
    }
    \subfloat[Examples by CDM score range]{
    \begin{minipage}{0.35\textwidth}
        \centering
        \includegraphics[width=\textwidth]{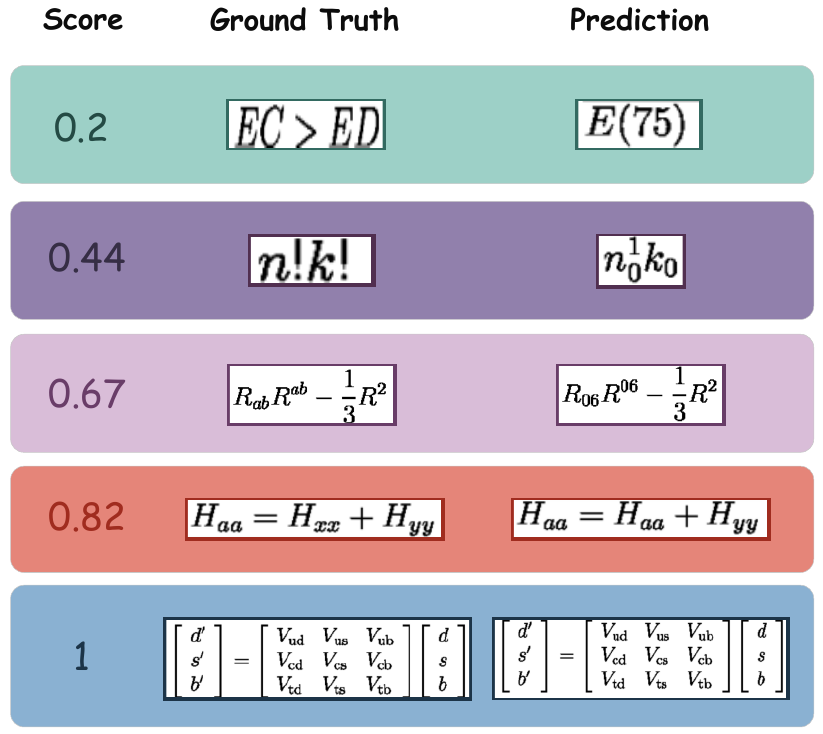}
        \label{fig:fig3_b}
    \end{minipage}
    }
\label{fig:fig3}
\vspace{-4pt}
\caption{CDM Score Distribution and Example Analysis. (a) CDM score distribution for Mathpix, UniMERNet, Texify, and Pix2Tex on the UniMER-Test dataset. (b) Example analysis of Pix2Tex predictions across different CDM score ranges.
}
\vspace{-15pt}
\end{figure*}

\subsection{Metric Calculation}

We use the F1-Score as the default metric for evaluating CDM (Character Detection Metric), defined as:
\begin{equation}
CDM  = \frac{2 \times TP}{2 \times TP + FP + FN},
\end{equation}
where $TP$ denotes true positives, $FP$ denotes false positives, and $FN$ denotes false negatives.

To further evaluate the accuracy of formula recognition, we introduce the \( ExpRate@CDM \) metric , defined as:
\begin{equation}
    ExpRate@CDM = \frac{\sum_{i=1}^{N} \mathbb{I}(CDM_i = 1)}{N},
\end{equation}
where $\mathbb{I}$ is the indicator function that equals 1 if $CDM_i = 1$ and 0 otherwise, and $N$  is the total number of formulas. This metric represents the proportion of formulas for which the model's prediction results are perfectly matched. Essentially, $ExpRate@CDM$ serves as a precise version of the ExpRate metric specifically for formula recognition.

\vspace{7pt}
\section{Experiments}

\begin{figure*}[h]
    \centering
    \subfloat[Human Evaluation Preferences]{
    \begin{minipage}{0.26\textwidth}
        \centering
        \includegraphics[width=\textwidth]{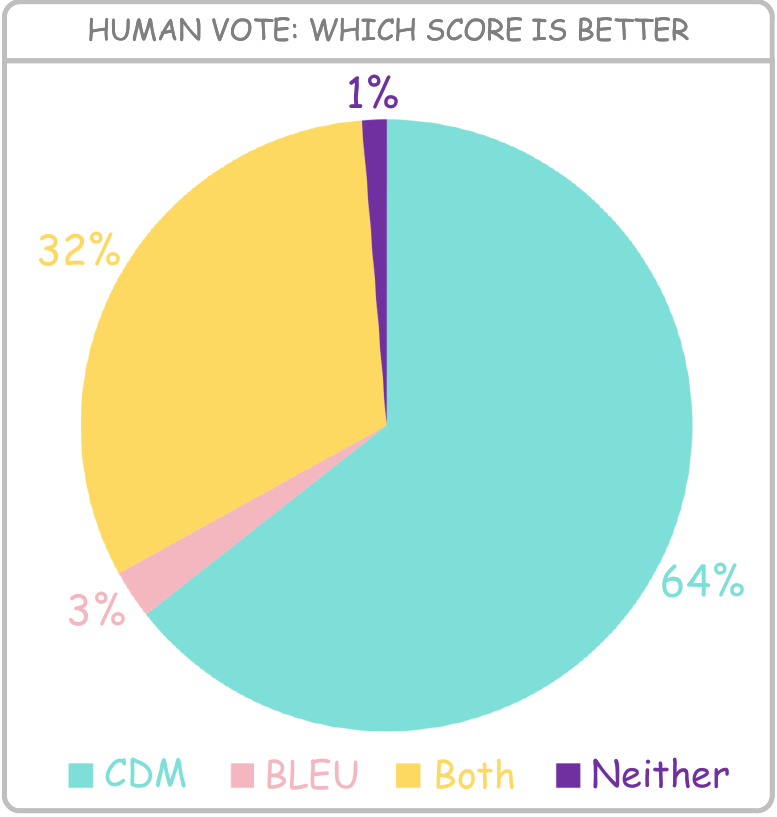}
        \label{fig:fig4_a}

    \end{minipage}
    }
    \subfloat[BLEU vs. CDM Score Distribution]{
    \begin{minipage}{0.37\textwidth}
        \centering
        \includegraphics[width=\textwidth]{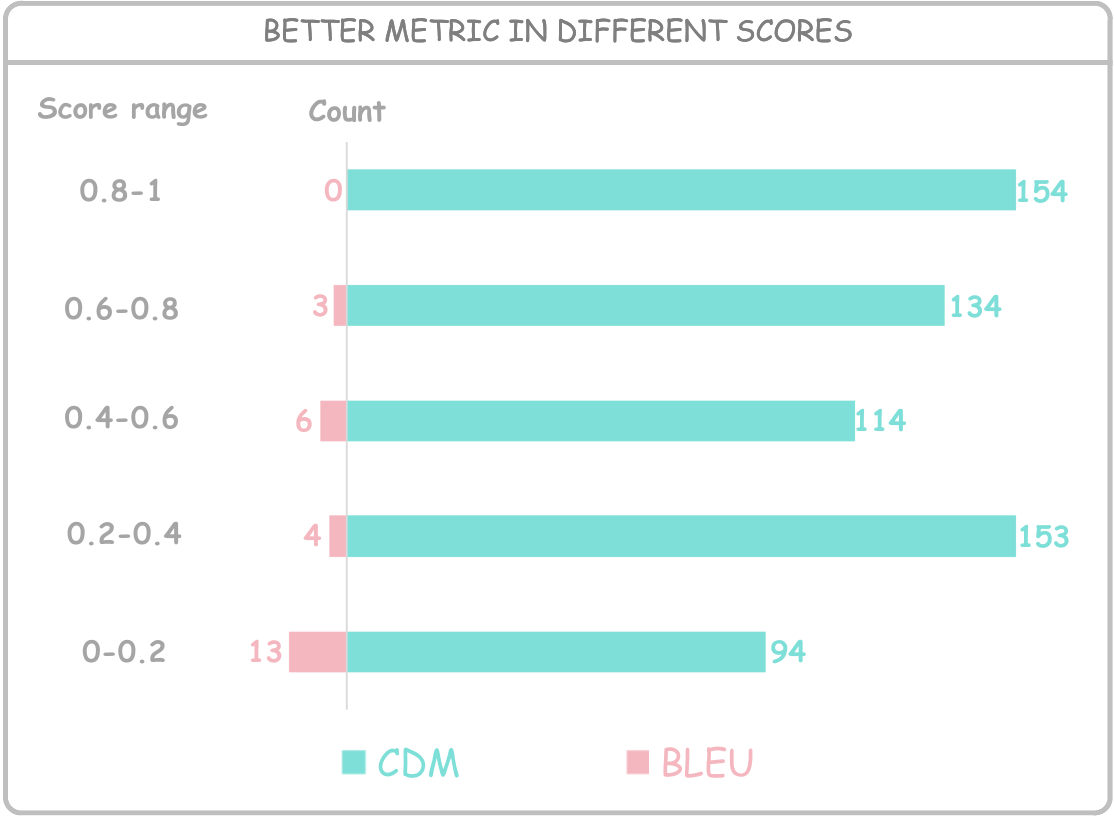}
        \label{fig:fig4_b}
    \end{minipage}
    }
    \subfloat[Impact of Formula Writing Styles]{
    \begin{minipage}{0.31\textwidth}
        \centering
        \includegraphics[width=\textwidth]{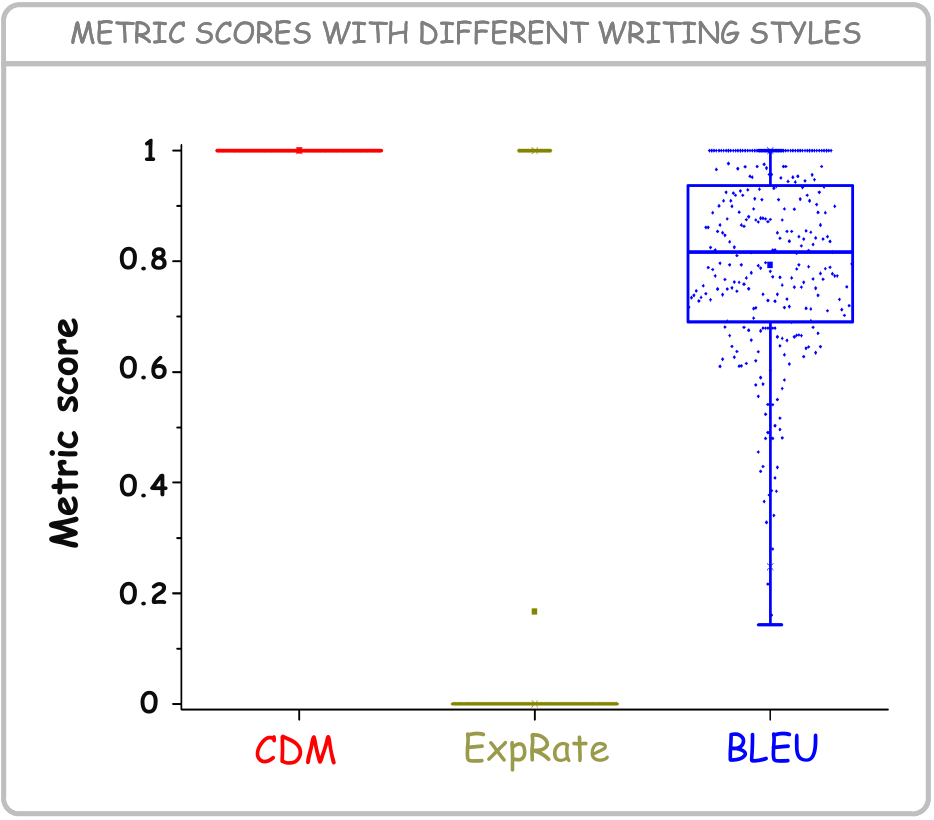}
        \label{fig:fig4_c}
    \end{minipage}
    }
\label{fig:fig4}
\vspace{-6pt}
\caption{Human evaluation and metric comparison. (a) Distribution of human preferences. (b) Distribution of cases preferring BLEU or CDM across different score ranges. (c) Impact of formula writing styles on BLEU, ExpRate and CDM scores.}
\vspace{-12pt}
\end{figure*}

\begin{figure}[ht]
\begin{center}
    \includegraphics[width=0.95 \linewidth]{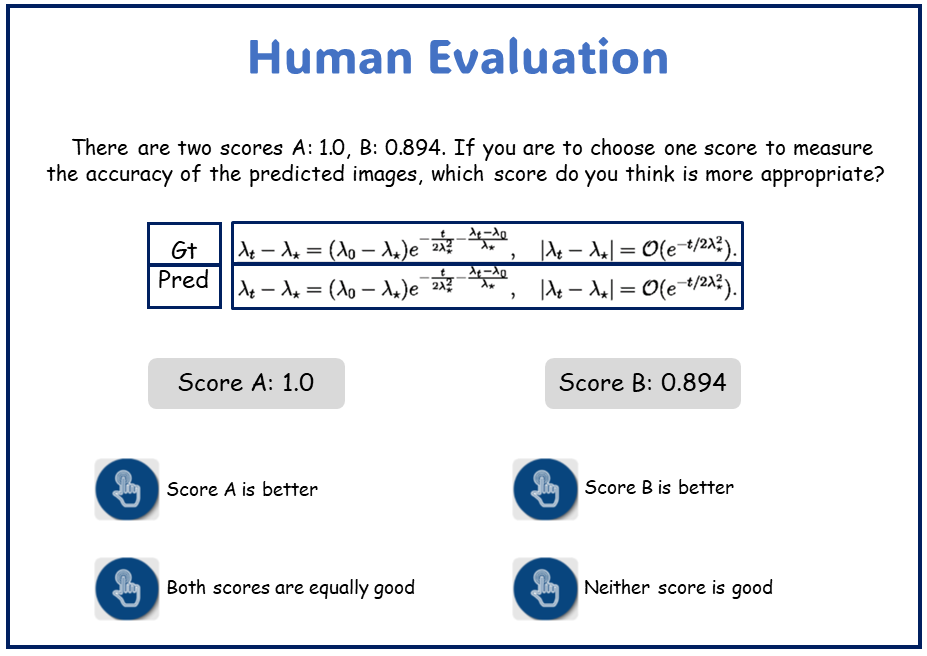}
\end{center}
\vspace{-8pt}
    \caption{Annotation interface for evaluating user preferences between CDM and BLEU scores. Annotators chose from four options: ``Score A is better", ``Score B is better", ``Both scores are equally good", or ``Neither score is good".}
\vspace{-10pt}
\label{fig:fig_GUI}
\end{figure}

\subsection{Models and Data}

We validate the CDM metric by evaluating several mainstream formula recognition models using both subjective impressions and objective metrics. The models include open-source UniMERNet~\cite{wang2024unimernet}, Texify~\citep{texify2023}, Pix2tex~\citep{pix2tex2022}, and the commercial Mathpix API, all tested on the UniMER-Test dataset. Besides, we evaluate document-level models, such as the open-source Nougat~\cite{blecher2023nougat} and the commercial GPT-4o~\cite{gpt4o}. Vary~\cite{wei2023vary} and StrucTexTv3~\cite{lyu2024structextv3} are excluded as they are currently unavailable.

\noindent \textbf{UniMER-Test Dataset.} The dataset includes 23,757 formula samples, categorized into Simple Printed Expressions (SPE), Complex Printed Expressions (CPE), Screenshot Expressions (SCE), and Handwritten Expressions (HWE). We use these categories to conduct the model evaluation.

\noindent \textbf{Tiny-Doc-Math Dataset.} To evaluate document-level recognition, we construct the Tiny-Doc-Math dataset, consisting of arXiv papers in mathematics and computer science, published after June 2024, to ensure that they are not in the training data of the compared models. We obtain LaTeX code and corresponding PDFs, match displayed equations using regular expressions, and manually verify them. Overall, the dataset includes 12 PDFs, totaling 196 pages and 437 formulas.

This validation set includes both formula-level and document-level evaluations:
\begin{itemize}
\item \textbf{Formula-level:} Using single rendered formula images as input, we evaluate Mathpix, Pix2Tex, and UniMERNet. These models accept cropped formula images as input, and we compare the model outputs with the ground truth to compute relevant metrics.
\item \textbf{Document-level:} Using PDFs or images as input, we evaluate Nougat, GPT-4o, and Mathpix, which can convert entire PDF pages into Markdown format. We match the displayed equations in the model outputs using regular expressions and compare them with the ground truth LaTeX formulas to compute relevant metrics.
\end{itemize}

\subsection{Credibility Assessment of CDM}

\subsubsection{Rendering Success Rate}

The CDM metric relies on the successful rendering of formula images. For models that fail to render images, we assign a CDM score of 0, as rendering failures indicate that the predicted LaTeX code lacks critical elements. The rendering success rates on the UniMER-Test dataset for Pix2tex, Texify, UniMERNet, and Mathpix are 96.63\%, 94.77\%, 99.71\%, and 97.82\%, respectively, ensuring the applicability and reliability of the CDM metric.

\subsubsection{User Preference Evaluation}

We analyze the distribution of CDM scores for four models on the UniMER-Test dataset. As shown in \Cref{fig:fig3_a}, Mathpix and UniMERNet perform well in terms of CDM scores. We conduct a detailed analysis of the Pix2Tex model by randomly selecting samples from different score ranges to evaluate if the prediction quality corresponds to the CDM scores. The analysis in \Cref{fig:fig3_b} shows that the CDM scores effectively reflect formula quality, with higher scores indicating fewer errors.

To verify the consistency between the CDM metric and human evaluation, we select 1,008 CDM scores from Pix2Tex predictions, ensuring a balanced score distribution. We design an annotation interface (shown in \Cref{fig:fig_GUI}), displaying a ground truth label and the corresponding predicted LaTeX rendered image. Annotators choose between ScoreA, ScoreB, Both (credible), and Neither (credible). ScoreA and ScoreB correspond to the BLEU and CDM scores, respectively, but their order is randomized. For more details, please refer to the supplementary materials.

\begin{table}[h]
    \resizebox{0.49\textwidth}{!}{
        \begin{tabular}{lcccc}
        \toprule
         \textbf{Model} & \textbf{ExpRate} & \textbf{ExpRate@CDM} & \textbf{BLEU} & \textbf{CDM} \\
         \midrule
        Pix2tex   &\underline{0.1237} & 0.2910 & 0.4080  & 0.6360 \\
        Texify    &\underline{0.2288} & 0.4950 & 0.5890  & 0.7550 \\      
         % \rowcolor{gray!20}
        Mathpix   &\underline{0.2610} & 0.5000 & 0.8067 & 0.9510 \\
        UniMERNet &\underline{0.4799} & 0.8110 & 0.8425 & 0.9680 \\
         \bottomrule
    \end{tabular}
    }
    \caption{UniMER-Test evaluation results. Comparison of mainstream models using different metrics.}
    \label{tab:tab1}
    \vspace{-15pt}
\end{table}

The results in \Cref{fig:fig4_a} show that 64\% of participants prefer the CDM metric, and 32\% consider both metrics good. This indicates a 96\% consistency between the CDM metric and human evaluation, demonstrating its reliability. \Cref{fig:fig4_b} compares the number of cases where BLEU or CDM is preferred across different score ranges, showing that CDM consistently outperforms BLEU.

\subsubsection{Objective Stability Assessment}

To evaluate the impact of formula writing styles on the CDM and BLEU metrics, we randomly select 50 formulas with LaTeX source code and rewrite each formula five times using GPT-4, generating 250 additional formulas. We manually verify these formulas to ensure their rendered results are identical to the original 50 formulas. Using the initial LaTeX source code as the ground truth, we analyze the score distribution of the BLEU and CDM metrics. As shown in \Cref{fig:fig4_c}, the CDM metric is unaffected by style changes, with all samples scoring 1. In contrast, the BLEU metric's scores are dispersed, making it unsuitable for formula evaluation. The CDM metric remains robust and reliable despite formatting changes.

\begin{table*}[ht]
\vspace{-8pt}
\centering
\resizebox{0.78\linewidth}{!}{
\setlength{\tabcolsep}{4pt}
{
\begin{tabular}{lcccccccc}
\toprule[.9pt]
\multirow{2}{*}{\textbf{Method}} & \multicolumn{2}{c}{\textbf{SPE}} & \multicolumn{2}{c}{\textbf{CPE}} & \multicolumn{2}{c}{\textbf{HWE}} & \multicolumn{2}{c}{\textbf{SCE}}  \\ 
\cmidrule(rl){2-3} \cmidrule(rl){4-5} \cmidrule(rl){6-7} \cmidrule(rl){8-9}  & BLEU $\uparrow$ & CDM $\uparrow$& BLEU $\uparrow$ & CDM $\uparrow$ & BLEU $\uparrow$ & CDM $\uparrow$ & BLEU $\uparrow$ & CDM $\uparrow$  \\  \midrule

Pix2tex~\cite{pix2tex2022}         & 0.8730 & 0.9619 & 0.6550 & 0.6489 & \underline{0.0120} & 0.2453 & \underline{0.0920} & 0.6762   \\ 
Texify~\cite{texify2023}           & 0.9060 & 0.9852 & 0.6900 & 0.7041 & 0.3410 & 0.5269 & 0.4200 & 0.7932   \\ 
Mathpix        & 0.7920 & 0.9729 & 0.8061 & 0.9671 & 0.8060 & 0.9318 & \underline{0.8182} & 0.9238   \\
UniMERNet~\cite{wang2024unimernet}       & 0.9170 & 0.9946 & 0.9160 & 0.9707 & 0.9210 & 0.9530 & \underline{0.6160} & 0.9461   \\ 
\bottomrule[.9pt]
\end{tabular}
}}
\vspace{-4pt}
\caption{UniMER-Test subset evaluation results. Analysis of anomalies in BLEU and CDM metrics across different subsets.}
% \vspace{-8pt}
\label{tab:tab2}
\end{table*}

\begin{table}[ht]
    \centering
    \resizebox{0.50\textwidth}{!}{
        \begin{tabular}{llccc}
        \toprule
         \textbf{Image Type} & \textbf{Model} & \textbf{BLEU} & \textbf{CDM} & \textbf{ExpRate@CDM} \\
         \midrule
        \multirow{4}{*}{\textbf{Formula}}  
        & Pix2tex   & 0.4648 & 0.74440 & 0.3684  \\
        & GPT-4o    & \underline{0.6431} & 0.7330 & 0.4324   \\      
        & UniMERNet & 0.6056 & 0.9396 & \textbf{0.6887}  \\
        & Mathpix   & 0.6112 & \textbf{0.9480} & 0.2105  \\
        \midrule
        \multirow{3}{*}{\textbf{Document}}  
        & GPT-4o    & 0.3411 & 0.6502 & 0.1670    \\      
        & Nougat    & 0.5897 & 0.8326 & 0.6086   \\
        & Mathpix   & 0.5939 & \textbf{0.9567} & \textbf{0.6292}   \\
        \bottomrule
    \end{tabular}
    }
    % \vspace{-6pt}
    \caption{Tiny-Doc-Math Evaluation Results. Performance of mainstream models on cropped formula inputs and document-level screenshots using BLEU and CDM metrics.}
    % \vspace{-6pt}
    \label{tab:tab3}
\end{table}

\subsection{Evaluation of Mainstream Models}

We conduct a detailed evaluation of mainstream models using both the CDM and BLEU metric. Note that all BLEU metric in this paper have been normalized ~\cite{deng2017image,pix2tex2022}. However, as discussed in the limitation section, normalization operations cannot address all issues, which will be evident in the following experiments.

\subsubsection{UniMER-Test Evaluation}

As shown in \Cref{tab:tab1}, the evaluation results of the four models on the UniMER-Test dataset indicate that the quality ranges from low to high as follows: Pix2Tex, Texify, Mathpix, and UniMERNet, based on both BLEU and CDM metrics. ExpRate@CDM clearly shows the proportion of completely correct predictions for each model, indicating that the text character-based ExpRate is unreliable.

From the results in \Cref{tab:tab1}, it appears that the trends of the BLEU and CDM metrics are consistent. To verify the reliability of using the BLEU metric for model comparison, we further present evaluation results on the UniMER-Test subsets. As shown in \Cref{tab:tab2}, we observe two notable anomalies: Firstly, in the SCE subset, when comparing the quality of the Mathpix and UniMERNet models, the BLEU and CDM metrics provide opposite conclusions. A detailed review of the UniMERNet paper reveals that the SCE subset was annotated based on Mathpix and then manually corrected. This means that the expression style of the SCE formulas is more consistent with Mathpix. Consequently, even though the CDM metric indicates that UniMERNet has better actual model quality, the BLEU metric, influenced by the expression style, suggests that Mathpix is superior. Secondly, for the Pix2Tex model, the BLEU metric is very low on the HWE and SCE subsets but performs well on the SPE and CPE subsets. This discrepancy arises because the Pix2Tex training set includes a large number of printed formulas from arXiv and lacks data in the HWE and SCE styles.

These anomalies clearly illustrate the limitations of the BLEU metric in evaluating the quality of formula recognition models. In contrast, the CDM metric proposed in this paper is fair and intuitive.

\subsubsection{Tiny-Doc-Math Evaluation}

The evaluation results of Tiny-Doc-Math are shown in \Cref{tab:tab3}. For cropped formula inputs (formula-level), all four models perform reasonably well, with CDM scores above 0.7. Notably, the current leading multimodal large model GPT-4o has the highest BLEU score among the four models but the lowest CDM score. This discrepancy indicates that the BLEU metric may not be reliable, suggesting that the formula recognition accuracy of GPT-4o still has room for improvement, lagging behind traditional SOTA models. Additionally, although Mathpix has the highest CDM score, only 21.05\% of the formulas are completely accurate. Manual verification revealed that many formulas are missing commas or periods at the end.

When the input is document-level screenshots, the models output the recognition results for the entire document (not just the formulas). Evaluation is conducted by matching the recognized block formulas. In this scenario, it can be observed that the accuracy of GPT-4o further decreases. In contrast, Mathpix and Nougat perform better, but even the document multimodal large model Nougat only achieves a CDM score of 0.8326. This indicates that there is still significant room for improvement in document-level recognition models. Mathpix remains the best performer, with a fully correct formula rate of 62.92\%. The accuracy of document-level recognition is crucial for advanced document understanding tasks like scientific knowledge Q\&A, and CDM provides an excellent standard for selecting formula models and offers direction for improving formula recognition.

\section{Conclusion}

In this paper, we introduced Character Detection Matching (CDM), a novel evaluation metric for formula recognition task. CDM addresses the shortcomings of the existing metrics in formula recognition by utilizing spatial character matching, overcoming the challenges posed by diverse formula representations. Comprehensive evaluations on different models and datasets demonstrate CDM's superiority in precisely reflecting recognition quality, showing more intuitive assessment and paving the way for future research and improvements in the formula recognition field.

{
    \small
    \bibliographystyle{ieeenat_fullname}
    \bibliography{cdm}
}

% WARNING: do not forget to delete the supplementary pages from your submission 
\clearpage
\setcounter{page}{1}
\maketitlesupplementary

\section{User Preference Evaluation Analysis}

To provide a more intuitive and clear analysis of the credibility of CDM, we supplement the content in Section 5.2 with a detailed examination of user preferences for CDM and BLEU metrics under different conditions.

To assess the reliability of CDM, we design an annotation interface as shown in \Cref{fig:fig_GUI}. Given the ground truth rendered image and the model's predicted rendered image for various samples, annotators are asked to assign an appropriate score. Score A and Score B correspond to the BLEU and CDM scores of the prediction results, but the order is randomized so that users do not know which score corresponds to which metric. Users make their choice based on their intuitive judgment from four options.

\begin{figure*}[ht]
\begin{center}
    \includegraphics[width=0.85 \linewidth]{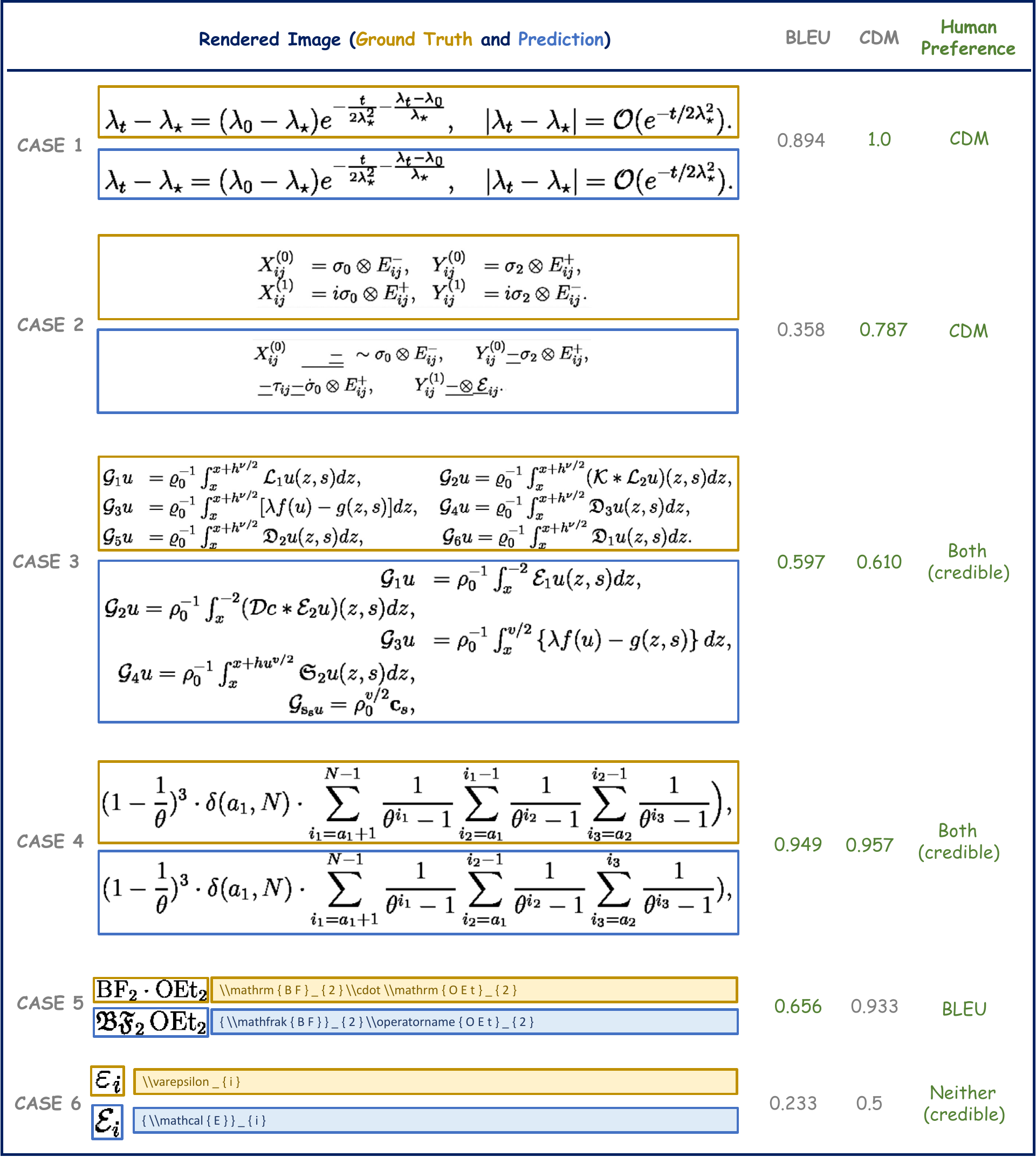}
\end{center}
    \caption{Examples of different human preferences (\textbf{CDM}, \textbf{BLEU}, \textbf{Both} (credible), \textbf{Neither} (credible)). Case 5 and Case 6 highlight some erroneous instances of CDM. In Case 5, CDM overlooks differences in character rendering styles, treating \texttt{"\textbackslash mathfrac\{BF\}"} as identical to \texttt{"\textbackslash mathrm\{BF\}"}, despite their visual differences. Conversely, in Case 6, CDM distinguishes between \texttt{"\textbackslash mathcal\{E\}"} and \texttt{"\textbackslash varepsilon"}, although they render similarly to human perception.}
\label{fig:fig_cdm_case}
\end{figure*}

A total of 1008 samples are scored, and the results are categorized into four scenarios. We provide a detailed and clear analysis of user preferences for CDM and BLEU metrics in each scenario, as illustrated in \Cref{fig:fig_cdm_case}:

\noindent \textbf{CDM is better (64\%)}: 
   In this scenario, examples include Case 1 and Case 2. In Case 1, the prediction result is 100\% correct, with a CDM score of 1 and a BLEU score of 0. Users directly chose the CDM score. In Case 2, the prediction result is mostly correct, but the BLEU score is significantly lower than expected, leading users to prefer the CDM score.

\noindent \textbf{Both scores are equally good (32\%)}:
   Examples in this scenario include cases 3 and 4, where the CDM and BLEU scores are relatively close, both reflecting the proportion of model prediction errors in an accurate and intuitive manner.

\noindent \textbf{BLEU is better (3\%)}:
   In Case 5, due to different token representations of "BF", BLEU detects inconsistencies, while CDM considers \(\mathrm{BF}\) and \(\mathfrak{BF}\) as the same token.

\noindent \textbf{Neither score is good (1\%)}:
   In Case 6, although the two formulas contain different tokens, \texttt{"\textbackslash mathcal\{E\}"} and \texttt{"\textbackslash varepsilon"}, they render similar images ($\mathcal{E}$ and $\varepsilon$). Both CDM and BLEU fail in this case.

CDM is reliable in 96\% of cases. The remaining 4\% are due to LaTeX issues, which will be optimized in future versions, with minimal impact on the overall evaluation.

\subsection{Latex Rendering and Syntax Errors}

CDM relies on normalizing LaTeX source code and rendering images. Therefore, code that cannot be rendered or contains syntax errors (which cannot be normalized) will result in computation failures. For example, the expression  \texttt{"z = \textbackslash left( \textbackslash begin\{array\}\{cc\}
 x \textbackslash \textbackslash ~y"} is a failure case due to a missing \texttt{"\textbackslash end\{array\}"}, leading to rendering failure. For these cases, CDM assigns a score of 0. Although CDM cannot directly handle them, this approach is reasonable and aligns well with human perception.
 
The number of LaTeX rendering and syntax errors depends on the quality of the model's prediction. Among the four models, Pix2tex, Texify, Mathpix, and UniMERNet, the proportion of LaTeX rendering and syntax errors in the predicted results on the UniMER-Test is 13.83\%, 5.03\%, 2.38\%, and 1.05\%, respectively.

\subsection{Rendering Types Affecting Token Consistency}

CDM defines characters without considering rendering styles. However, different rendering styles can produce visually distinct results, potentially causing different tokens to render into nearly identical characters(\Cref{fig:fig_cdm_case} Case6), or same tokens to render into different characters(\Cref{fig:fig_cdm_case} Case5). Similar situations include \texttt{"G"} and \texttt{"\textbackslash mathcal \{ G \}"}, \texttt{"\textbackslash mathcal \{ X \}"} and \texttt{"\textbackslash mathfrac \{ X \}"}, whose rendering effects are $G, \mathcal{G}, \mathcal{X}, \mathfrak{X}$, respectively. This inconsistency can confuse the token consistency check, leading to errors in the model's output.

\begin{figure*}[ht]
\begin{center}
    \includegraphics[width=0.92 \linewidth]{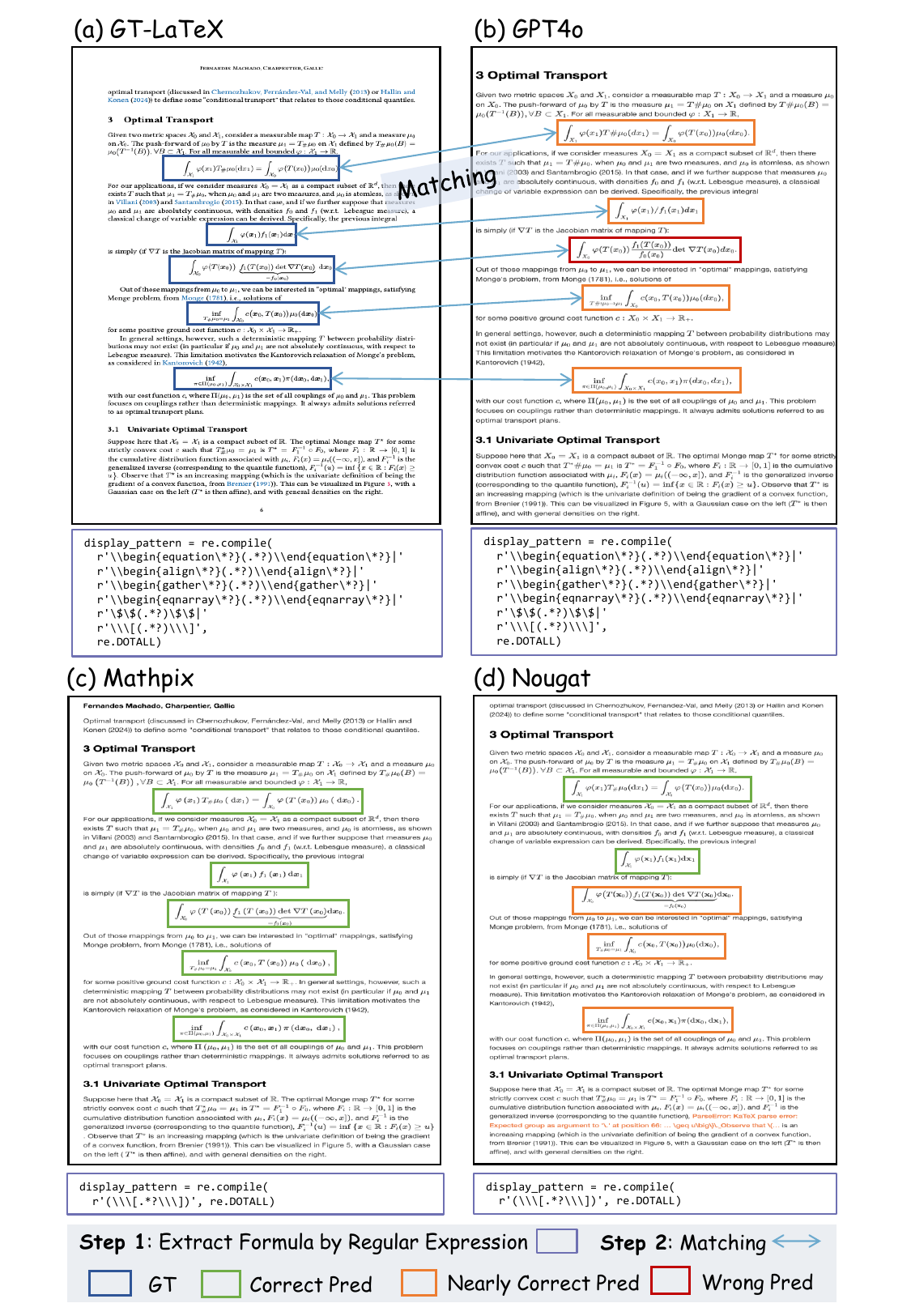}
\end{center}
    \caption{Detailed Process of Document-Level Evaluation.}
\label{fig:figA3}
\end{figure*}

\begin{figure*}[ht]
\begin{center}
    \includegraphics[width=0.8 \linewidth]{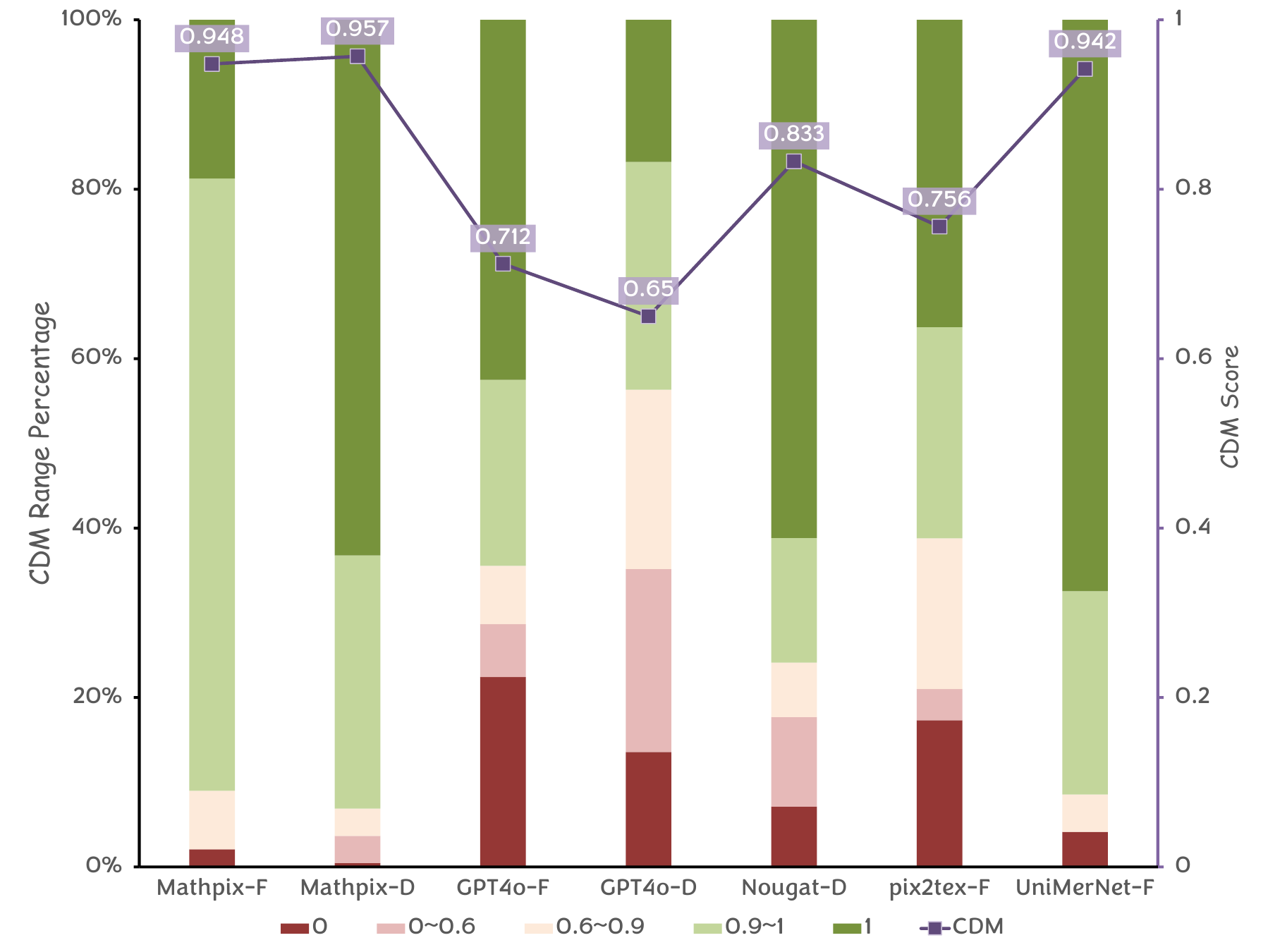}
\end{center}
    \caption{The CDM range percentage and CDM score of each model. ``F" indicate Formula-Level, ``D" indicate Document-Level.}
\label{fig:figA4}
\end{figure*}

\begin{figure*}[ht]
\begin{center}
    \includegraphics[width=0.9 \linewidth]{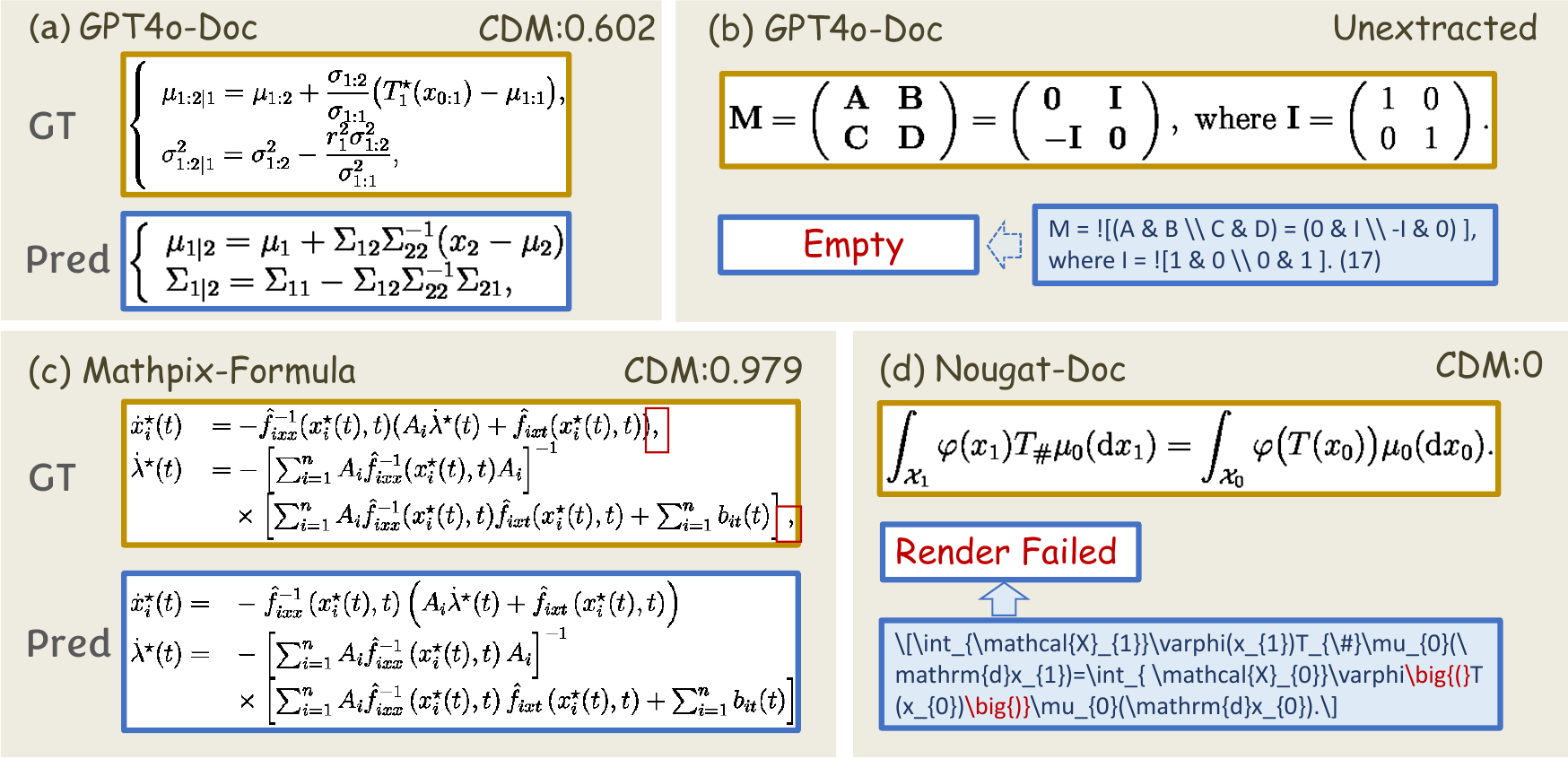}
\end{center}
    \caption{Examples of common prediction errors in GPT-4o, Mathpix, and Nougat.}
\label{fig:figA5}
\end{figure*}

\section{In-Depth Methodology for Evaluating Tiny-Doc-Math}

\subsection{Construction of Tiny-Doc-Math Dataset}
The evaluation dataset is constructed primarily from arXiv papers in the fields of mathematics and computer science, published after June 2024. We manually select a batch of these papers and download the LaTeX source code and corresponding PDFs. Using regular expressions, we match the formulas displayed from the LaTeX source. After individual formula rendering and manual verification, the Tiny-Doc-Math validation set is built, comprising 12 papers, 196 pages, and a total of 437 formulas.

\subsection{Formula-Level Evaluation Methodology}
Once the evaluation dataset is constructed, we extract mathematical formulas from the LaTeX source code. Since LaTeX sources may contain custom commands and comments from authors, we apply a series of preprocessing steps to ensure accurate extraction. First, we remove comments from the LaTeX source using regular expressions (including \texttt{"\%"}, \texttt{"\textbackslash iffalse... \textbackslash fi"}, and \texttt{"\textbackslash begin\{comment\}...\textbackslash end\{comment\}"}). Next, we convert aliases defined by commands such as \texttt{"\textbackslash newcommand\{\}\{\}"}, \texttt{"\textbackslash renewcommand\{\}\{\}"}, \texttt{"\textbackslash DeclareMathOperator\{\}\{\}"}, \texttt{"\textbackslash DeclareMathOperator*\{\}\{\}"}, \texttt{"\textbackslash def\textbackslash...\{\}{}"}, and \texttt{"\textbackslash DeclareRobustCommand\{\}\{\}"} to their original forms to ensure successful formula rendering. We then remove content before \texttt{"\textbackslash begin\{document\}"} to avoid matching irrelevant information. After preprocessing, we extract displayed mathematical formulas from the LaTeX source using a series of regular expressions, as shown in \Cref{fig:figA3}(a). For each paper, the matched mathematical formulas are written to a text file, one formula per line.

We render the extracted GT mathematical formulas to obtain formula-level GT images, which are then used as inputs for Mathpix, UniMerNet, pix2tex, and GPT-4o to generate corresponding predictions. Finally, we compute metrics such as BLEU and CDM after matching the predictions with the GTs.

\subsection{Document-Level Evaluation Methodology}
We convert PDF pages to images and use these images as inputs for Mathpix and GPT-4o to generate corresponding predictions, while Nougat takes the whole PDF as input. After obtaining the document-level predictions, we used extraction algorithms to extract displayed formulas from the predictions, and match them with the GT formulas obtained in the previous section to compute BLEU and CDM metrics. 

Due to the different syntax formats of the outputs from different models, we use different regular expressions to extract formulas for each model, as shown in \Cref{fig:figA3}(b), (c), and (d). Similarly, for each PDF, the matched mathematical formulas from each model's predictions are written to a text file, one formula per line.

\subsection{Matching and Metric Computation}
After obtaining the GTs and predicted mathematical formulas, we match the GTs with the predicted formulas line by line to compute the final CDM metric. Given the high accuracy of displayed formula predictions, we use edit distance as the metric for matching formulas. To account for different math delimiters used by different models (\textit{e.g.}, \texttt{"\textbackslash begin\{equation\}...\textbackslash end\{equation\}"} vs. \texttt{"\textbackslash[...\textbackslash]"} ), we remove all math delimiters before matching, focusing solely on the content. Labels and tags are also removed from the formulas.

The matching process consists of two rounds. In the first round, we set a low edit distance threshold for precise matching. This means that only predictions with a high similarity to the ground truth formula will be matched. We iterate through the GT formulas, calculating the edit distance with all predicted results. The prediction with a minimum edit distance is recorded as matched only if the minimum edit distance was below the threshold. If not, we skip the line and mark both the GT and the prediction as unmatched. In the second round, we set a higher threshold to account for those matching cases where the edit distance might be large. We iterate through the unmatched GT formulas, calculate the edit distance with the remaining unmatched predicted formulas, and record matches if the distance is below the threshold. If any predicted formulas remain unmatched after the first two rounds, we mark them as incorrect or redundant predictions and append them to the end of the matched results.

Through practical implementation, we find that setting the first-round threshold to 0.4 and the second-round threshold to 0.8 provides the most reasonable matching. Although extreme cases might occur where the rendered results are identical but fail to match due to large edit distances, these instances are not common and have been manually corrected. 

After matching the GTs and predicted formulas, we compute metrics such as BLEU and CDM.

\subsection{Result Discussion}
As shown in \Cref{fig:figA4}, GPT-4o's document-level predictions exhibited a significant number of CDM scores between 0.6 and 0.9, primarily due to hallucination phenomena in large models. For example, as shown in \Cref{fig:figA5}(a), GPT-4o generates structurally similar but content-irrelevant results. Additionally, as shown in \Cref{fig:figA5}(b), GPT-4o's predictions often lack standardized formatting, \textit{i.e.}, frequently generating formulas without math delimiters, leading to extraction and rendering failures and resulting in many CDM=0 cases. For Mathpix, although the CDM between the document level and formula level is close, the proportion of CDM=1 predictions at the formula level is significantly lower. This is mainly due to the lack of commas in Mathpix's single formula predictions, as shown in \Cref{fig:figA5}(c). Nougat's predictions often contain syntax errors, as shown in \Cref{fig:figA5}(d), leading to rendering failures and CDM=0 cases. Moreover, Nougat's predictions sometimes leave several pages in the middle of the PDF with no prediction results, resulting in missing formulas in the final output.

\clearpage

\begin{figure*}[t]
\begin{center}
    \includegraphics[width=0.8 \linewidth]{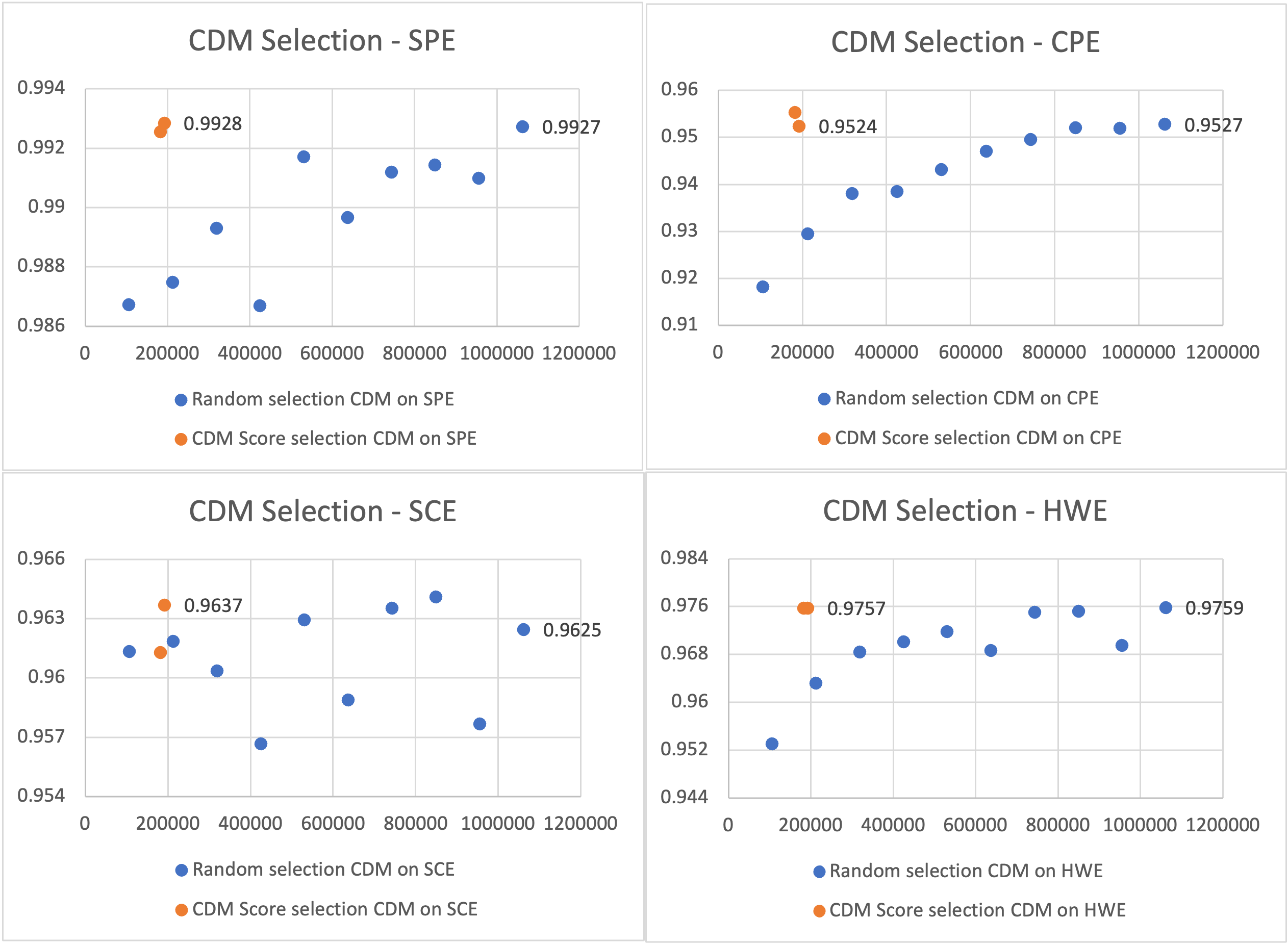}
\end{center}
    \caption{CDM metrics on four UniMER-Test subsets (SPE, CPE, SCE, HWE) for models trained with varying amounts of data (10\%, 20\%, ..., 100\%) and models trained using two rounds of hard case selection. The scatter plot shows performance improvements with increasing training data and the efficiency of hard case selection.}
\label{fig:fig6_data_select}
\end{figure*}

\begin{figure*}[ht]
\begin{center}
    \includegraphics[width=0.8 \linewidth]{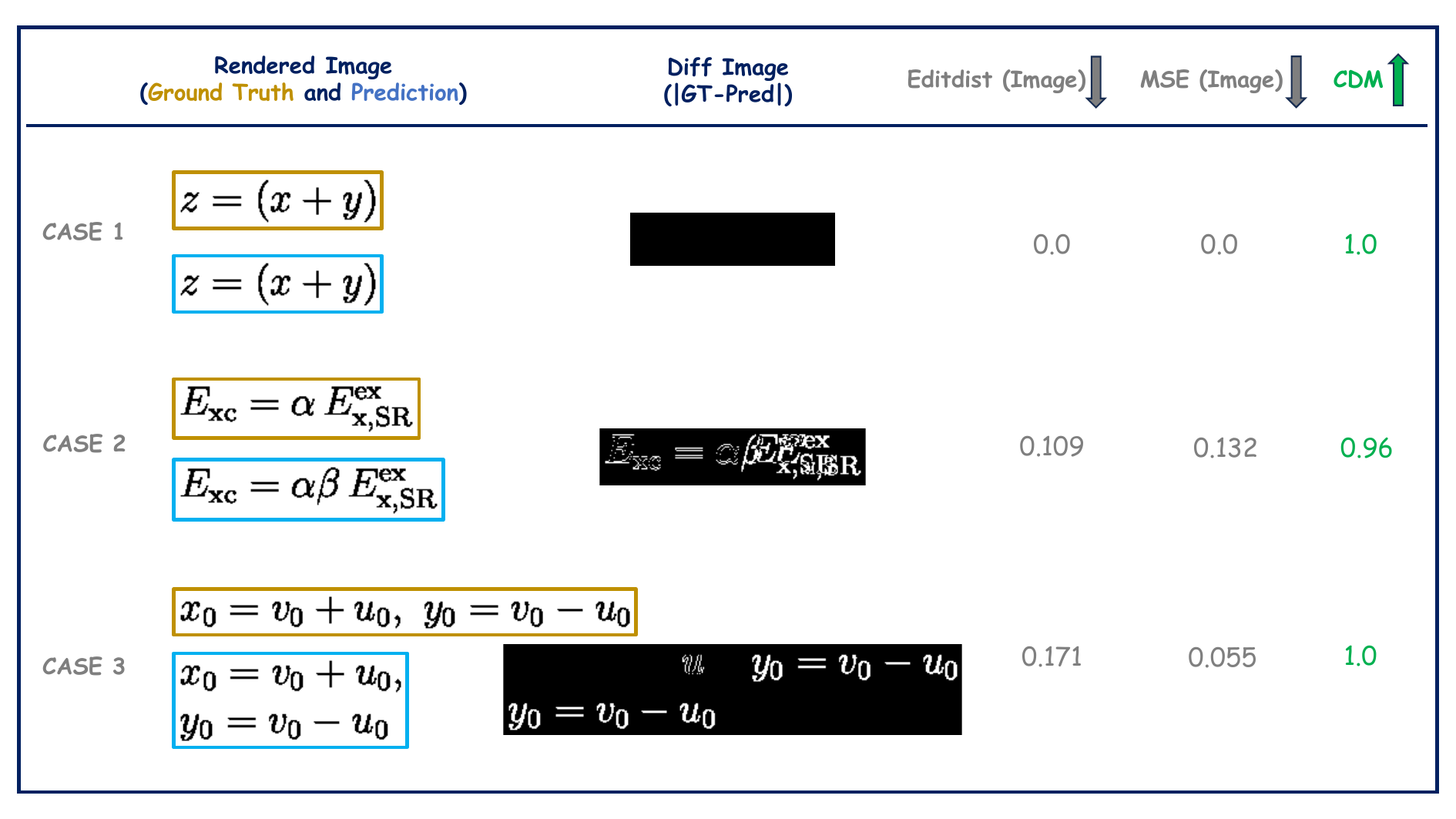}
\end{center}
    \caption{Examples of formula recognition evaluation using image edit distance and MSE. Case 1: Correct prediction with zero Editdist and MSE. Case 2: Missing one \(\alpha\) causing all subsequent positions to mismatch. Case 3: Correct formula content but an extra newline character causing significant image difference.}
\label{fig:fig7_image_metric}
\end{figure*}

\clearpage

\section{Efficient Data Selection for Formula Recognition}

% \begin{table}[t]
%     \centering
%     \resizebox{0.48\textwidth}{!}{
%         \begin{tabular}{llccc}
%         \toprule
%          \textbf{Image Type} & \textbf{Model} & \textbf{BLEU} & \textbf{CDM} & \textbf{ExpRate@CDM} \\
%          \midrule
%         \multirow{4}{*}{\textbf{Formula}}  
%         & Pix2tex   & 0.4648 & 0.74440 & 0.3684  \\
%         & GPT-4o    & \underline{0.6431} & 0.7330 & 0.4324   \\      
%         & UniMERNet & 0.6056 & 0.9396 & \textbf{0.6887}  \\
%         & Mathpix   & 0.6112 & \textbf{0.9480} & 0.2105  \\
%         \midrule
%         \multirow{3}{*}{\textbf{Document}}  
%         & GPT-4o    & 0.3411 & 0.6502 & 0.1670    \\      
%         & Nougat    & 0.5897 & 0.8326 & 0.6086   \\
%         & Mathpix   & 0.5939 & \textbf{0.9567} & \textbf{0.6292}   \\
%         \bottomrule
%     \end{tabular}
%     }
%     % \vspace{-3pt}
%     \caption{Tiny-Doc-Math Evaluation Results. Performance of mainstream models on cropped formula inputs and document-level screenshots using BLEU and CDM metrics.}
%     % \vspace{-7pt}
%     \label{tab:tab3}
% \end{table}

\begin{table*}[t]
\vspace{-10pt}
    \centering
    \resizebox{0.95\linewidth}{!}{
    \begin{tabular}{lcccccccc}
    \toprule
    \multirow{2}{*}{\textbf{Method}} & \multicolumn{2}{c}{\textbf{SPE}} & \multicolumn{2}{c}{\textbf{CPE}} & \multicolumn{2}{c}{\textbf{HWE}} & \multicolumn{2}{c}{\textbf{SCE}}  \\ 
    \cmidrule(rl){2-3} \cmidrule(rl){4-5} \cmidrule(rl){6-7} \cmidrule(rl){8-9}  & CDM $\uparrow$ & ExpRate@CDM $\uparrow$& CDM $\uparrow$ & ExpRate@CDM $\uparrow$ & CDM $\uparrow$ & ExpRate@CDM $\uparrow$ & CDM $\uparrow$ & ExpRate@CDM $\uparrow$  \\ 
    
    \midrule
    
    Pix2tex         & 0.9619  & 0.7240 & 0.6489 & 0.0705 & 0.2453 & 0.0060 & 0.6762 & 0.3284  \\ 
    Texify          & 0.9852 & 0.9104 & 0.7041 & 0.2821 & 0.5269 & 0.2359 & 0.7932 & 0.5132  \\ 
    Mathpix        & 0.9729 & 0.4400 & \textbf{0.9671} & 0.288 & 0.9318 & 0.5928 & 0.9238 & 0.7233   \\
    
    \midrule
    
    UniMERNet-tiny  & \underline{0.9910} & 0.9232 & 0.9491 & 0.6988 & 0.9328 & 0.6186 & \underline{0.9384} & 0.7655   \\
    UniMERNet-small  & 0.9906 & \textbf{0.9335} & 0.9588 & \underline{0.7767} & \underline{0.9370} & \underline{0.6393} & \textbf{0.9406} & \underline{0.7693}   \\
    UniMERNet-base  & \textbf{0.9914} & \underline{0.9329} & \underline{0.9595} & \textbf{0.8046} & \textbf{0.9400} & \textbf{0.6431} & 0.9373 & \textbf{0.7697}   \\
    
    \bottomrule
    \end{tabular}
    }
    % \vspace{-2pt}
    \caption{Newest UniMER-Test subset evaluation results}
    % \vspace{-10pt}
    \label{tab:tab4}
\end{table*}

Current formula recognition methods often overlook the importance of sample selection during training. We demonstrate that by utilizing the CDM metric for training data selection, it is possible to achieve performance comparable to using the entire dataset while only utilizing less than 20\% of the data. We conduct the following experiment: First, we randomly split the UniMER-1M dataset into ten equal parts. We then train the model using 10\%, 20\%, up to 100\% of the data and observe the model's performance with varying amounts of training data. As shown by the blue points in \Cref{fig:fig6_data_select}, the model's performance generally improves as the amount of training data increases. Notably, with just 10\% (106,179 samples) of the data, the model achieves satisfactory performance, accurately predicting most formulas. This suggests that the remaining 90\% of the data may be largely redundant for training purposes.

To further investigate, we perform two rounds of hard case data selection. First, we use the model trained on 10\% of the data to identify samples with CDM $\neq 1$ from the remaining 90\%. We find 76,026 such samples, which is less than 8\% of the remaining data, indicating that over 90\% of the formulas can be accurately predicted. Combining these with the initial 10\% random data, we have a total of 182,205 samples (17.16\% of the UniMER-1M dataset). As shown in \Cref{fig:fig6_data_select}, the model trained on this combined dataset performs comparably to the model trained on the full dataset, except for a slight underperformance on the SCE subset.

Next, we use this model to further select hard cases from the remaining data, identifying an additional 9,734 samples, representing about 1\% of the remaining data. This brings the total to 191,939 samples (18.08\% of the full dataset). The performance of this model shows a slight improvement over the previous round, achieving results comparable to or even exceeding those of the model trained on the full dataset across various subsets.

This experiment demonstrates the effectiveness of using CDM for hard case selection in formula recognition. Training based on hard case mining can serve as an efficient method to enhance model performance. This approach allows for the expansion of training data by selecting only the necessary samples, eliminating the need to use the entire dataset. Future formula recognition datasets can be expanded using this method, focusing on the most challenging samples to improve model accuracy and efficiency.

\section{Evaluation Method Based on Image Differences}

Previous work~\cite{wang2021translating} mentions using image-based difference methods for evaluating formula recognition results, but a thorough analysis of the limitations of this approach is needed. To further assess the effectiveness of these methods, we conduct experiments using both image edit distance (Editdist) and Mean Squared Error (MSE) of image differences. As shown in \Cref{fig:fig7_image_metric}, Case 1 demonstrates that when the model's prediction is correct and the rendered output perfectly matches the ground truth (GT), both EditDist and MSE are zero, indicating an accurate formula. However, in Case 2, where the prediction misses the character $\alpha$, the image-based difference method flags all subsequent positions as mismatched, even though only one character is missing. A more severe example is illustrated in Case 3, where the predicted formula content is correct but an extra newline character is predicted, leading to a significant image difference. In this case, both EditDist and MSE are non-zero and fail to reflect the error accurately. This highlights the necessity of the proposed CDM metric.

\section{Latest UniMERNet performance}

\Cref{tab:tab2} shows how UniMERNet~\cite{wang2024unimernet} compares to other models. It was recently updated with three model weights of different sizes, which we re-evaluated, and the results are shown in \Cref{tab:tab4}.

\end{document}